\documentclass{article}
\usepackage{kr}

\usepackage{times}
\usepackage{color, soul}
\renewcommand\hl[1]{#1}
\usepackage{url}
\usepackage[hidelinks]{hyperref}
\usepackage[utf8]{inputenc}
\usepackage[small]{caption}
\usepackage{graphicx}
\usepackage{amsmath}
\usepackage{amsthm}
\usepackage{booktabs}
\usepackage{algorithm}
\usepackage{algorithmic}
\usepackage{amsfonts}
\usepackage{subcaption}

\title{S-REINFORCE: A Neuro-Symbolic Policy Gradient Approach for Interpretable Reinforcement Learning}
\author{
Rajdeep Dutta$^{1*}$\and
Qincheng Wang$^{1,2}$\and
Ankur Singh$^{1 \dag}$\and
Dhruv Kumarjiguda$^{1,2 \dag}$\and
Li Xiaoli$^1$\And
Senthilnath Jayavelu$^{1*}$
\affiliations
$^1$Institute for Infocomm Research (I$^{2}$R), A*STAR, Singapore.\\
$^2$Nanyang Technological University, Singapore.\\
$\dag$ Equally contributed\\
 \emails $^*$ Corresponding authors:
rajdeep\_dutta@i2r.a-star.edu.sg,
j\_senthilnath@i2r.a-star.edu.sg
}

\begin{document}

\maketitle

\begin{abstract}
This paper presents a novel RL algorithm, S-REINFORCE, which is designed to generate interpretable policies for dynamic decision-making tasks. The proposed algorithm leverages two types of function approximators, namely Neural Network (NN) and Symbolic Regressor (SR), to produce numerical and symbolic policies, respectively. The NN component learns to generate a numerical probability distribution over the possible actions using a policy gradient, while the SR component captures the functional form that relates the associated states with the action probabilities. The SR-generated policy expressions are then utilized through importance sampling to improve the rewards received during the learning process. We have tested the proposed S-REINFORCE algorithm on various dynamic decision-making problems with low and high dimensional action spaces, and the results demonstrate its effectiveness and impact in achieving interpretable solutions. By leveraging the strengths of both NN and SR, S-REINFORCE produces policies that are not only well-performing but also easy to interpret, making it an ideal choice for real-world applications where transparency and causality are crucial.
\end{abstract}

\section{Introduction}
In recent years, reinforcement learning (RL) algorithms have gained significant research interest due to their effectiveness in finding optimal solutions to dynamic decision-making and control problems. 
Such solutions have shown tremendous success in various fields ranging from two-player games, robotics to medicine and drug discovery \cite{gamesRL_deepMind_Science2018,surveyRL_ieee2017,denovoRL_Science2018,chatGPT2022}.   
To solve sequential decision-making tasks involving continuous-valued state and action variables, function approximators in RL help to represent a value function determining the quality of state-action pairs and/or a policy mapping of the probabilities distributed over all possible actions. 

In the machine learning community, neural networks (NNs) have become popular due to their power of approximating complex, nonlinear and unknown functions in various applications \cite{1KR2022}. 
However, \hl{there is a lack of physical insight in the input-output mapping learned by an NN}, 
and it involves many non-linear operators and transformations that make the NN-generated policy difficult to deploy in real-world applications \cite{SymbPolicy_icml2021}. 
In contrast, symbolic regressors (SRs) offer interpretable input-output mappings by leveraging symbolic basis function expansions in their approximations \cite{rsos_2022} and they do not require a large amount of data for fitting purposes. 
To find analytical value functions, there exist offline methods such as: symbolic value iteration, symbolic policy iteration, and a direct solution to the Bellman equation \cite{Babuska_ieeeAccs2021}. The symbolic value function approximations were compact, mathematically tractable, and easily deployable. 
Also, in simulated and experimental environments, these symbolic value functions generated more stable control policies than NN \cite{Babuska_cdc2016}, \cite{SymbPolicy_icml2021}. 
Symbolic policies are readily interpretable, highly transparent, and easily reproducible due to their functional forms. Moreover, such policies can offer economic deployment solutions while satisfying memory or latency-related constraints \cite{SymbPolicy_icml2021}.    

Although SR produces analytical functions, the regression process involves a higher computational budget than training NN.
This is because when functions are encoded by strings of symbols, the number of such strings grows exponentially with string length \cite{AIFeynman_2020,KR2022}. 
Due to the combinatorial challenge of an exponentially large search space, symbolic regression remains an NP-hard machine learning problem \cite{transformerSR_icml2021,guiSR_ijcai2021,KR2022}.  
To maintain a balance between the interpretability of solutions and the associated computational cost, in this work, we leverage the strengths of both NN and SR into an RL framework. 
\hl{An NN is trained exhaustively to learn a numerical probability distribution over the possible actions, while an SR is fitted at regular intervals along the NN-training episodes to capture functional relations between the underlying states and the action probabilities.}
The contributions of this work are highlighted as follows:
\begin{itemize}
\item  To the best of our knowledge, \hl{this is the first attempt to generate symbolic policies in a cost-effective manner by combining the strengths of NN and SR in RL.}
\item  \hl{The main novelty lies in extracting interpretable policy expressions, and the performance improvement of an RL agent with importance sampling is a subsidiary.}
\item During the learning process, the \hl{periodic} knowledge transfer from NN to SR \hl{alleviates} the cost of symbolic regression throughout all training episodes.
\item Both NN and SR are trained concurrently in the proposed approach, and on completion of the training, any of the two learnt approximators can be used. 
\end{itemize}
Experimental results demonstrate that our proposed S-REINFORCE algorithm outperforms the REINFORCE algorithm in terms of interpretability and performance. 


\section{Related Works}  
Previous researchers have exploited genetic programming (GP) to symbolically approximate a policy obtained through classical control methods. However, this approach requires access to the governing dynamics equations that are not available in complex environments \cite{Babuska_ieeeAccs2021}.
To symbolically approximate a policy without explicitly knowing the underlying state-transition laws, model distillation and regression-based approaches have been utilized recently \cite{Hein2018}. 
Nevertheless, there lies an objective mismatch issue due to a conflict between the training objective of mimicking a pre-trained offline policy and the evaluation metric for improving an RL agent's performance.  
Rather, a direct search for symbolic policies has no objective mismatch conflict, wherein an autoregressive recurrent neural network (AutoRNN) is used to approximate a distribution over a discrete sequence of tokens representing operators, input variables and constants \cite{SymbPolicy_icml2021}.
A risk-seeking policy gradient algorithm can get along with the AutoRNN as it optimizes rewards for the best-case performance. 
The RNN-based risk-seeking policy gradient approach enables searching in the combinatorial space of expression trees, though it suffers from a limited exploration due to an early commitment phenomenon and/or initialization bias. 
On the other hand, a GP approach induces an exploration of the search space by means of a population of evolving candidate programs (trees), which involves high computational overhead.  
In practice, it is difficult to maintain a balance between the search space exploration and the associated computational cost in symbolic policy approximation procedures.  

Towards this direction, a recent work has motivated the use of two different approximators trained concurrently in RL using an online knowledge transfer mechanism. 
The proxy policy can be utilized via importance sample to reduce the variance of the gradient estimated with the main policy \cite{aamas_2019}.  
Therefore, instead of learning with data generated by a pre-trained offline policy, in this work, we train an SR using an online policy approximated by an NN. 
In our proposed framework, an NN is trained throughout all training episodes of RL, while an SR is fitted only at certain intervals. 
Thus, for symbolic policy approximation, we leverage the exploration power of a population-based genetic programming approach on a reasonable computational budget.  
Further, by using the numerically and the symbolically approximated policies, an importance sampling is applied for variance reduction of the gradient estimate.

\section{Model-free Policy Gradient}\label{sec:RL}
The present work focuses on the Markov Decision Process (MDP) type of sequential decision-making tasks. At time $t$, the agent observes its current state $s_t$ and takes an action $a_t$, which moves it to the next state $s_{t+1}$ with a reward $r_{t+1}$.     
Typically, an MDP is defined by a tuple of information: $<\mathbf{S}, \mathbf{A}, \mathbf{P}, R, \gamma, T>$, where $\mathbf{S}$ denotes a set of states; $\mathbf{A}$ is a set of actions; $\mathbf{P}=\{p(s'=s_{t+1}, r'=r_{t+1}|s=s_t, a=a_t)\}$ represents a set of transition probabilities; $R$ is the reward function; $0< \gamma< 1$ is the discount factor; and $T$ is the horizon of a trajectory $\tau=\{s_0,a_0,s_1,r_1,....,s_{T-1},a_{T-1},s_T,r_T\}$ generated by following a policy. A policy $\pi$ is a mapping from the set of states $\mathbf{S}$ to a probability distribution over the set of actions $\mathbf{A}$, and $\rho^\pi(s)$ is the probability of being in state $s$ while following policy $\pi$.     

\noindent The return expected from a trajectory $\tau$ (of horizon $T$) generated by following a parameterized policy $\pi_\theta$, is given by 
\begin{equation}\label{eqn:obj0_ExpRt}
    J(\theta) = \mathbf{E}_{\tau \sim \pi_\theta} [R(\tau)]~;~R(\tau)=\sum_{t=0}^{T-1}\gamma^t r_{t+1}.
\end{equation}

The policy gradient methods try to maximize the objective function in Equation (\ref{eqn:obj0_ExpRt}) by updating the policy parameters along the gradient ascent direction of it. The policy gradient is evaluated by 
\begin{eqnarray}\nonumber
&    \nabla_\theta J(\theta) = \nabla_\theta \mathbf{E}_{\tau \sim \pi_\theta} [R(\tau)] = \int_{\tau} \nabla_\theta \pi_\theta(\tau) R(\tau) d\tau\\\label{eqn:form0_PG}
&    or,~\nabla_\theta J(\theta) = \mathbf{E}_{\tau \sim \pi_\theta} [R(\tau) \nabla_\theta \log \pi_\theta(\tau)]~.
\end{eqnarray}
Since a new action-probability does not depend on the previous one according to Markov's principle of causality, $\pi_\theta(\tau)$ in the basic form of the policy gradient (\ref{eqn:form0_PG}) can be expanded with the product rule of probabilities as: $\pi_\theta(\tau)=\mathcal{P}(s_0) \prod \limits_{t=0}^{T-1} \pi_\theta(a_t|s_t) p(s_{t+1}, r_{t+1}|s_t, a_t)$, where the probabilities $\mathcal{P}(.)$ and $p(.|.)$ are inherent with a dynamic process \cite{lecture_algosRL}. Therefore, the derivative of $\log \pi_\theta(\tau)$ is:  
\begin{eqnarray}\nonumber
& \nabla_\theta \log \pi_\theta(\tau) = \nabla_\theta \{ \log \mathcal{P}(s_0) +\\ \label{eqn:expandPG}
& \sum \limits_{t=0}^{T-1} \log \pi_\theta(a_t|s_t) + \sum \limits_{t=0}^{T-1} \log p(s_{t+1}, r_{t+1}|s_t, a_t) \} 
\end{eqnarray}
\noindent The first and the third terms on the right hand side of Equation (\ref{eqn:expandPG}) are zero as the respective terms, $\log \mathcal{P}(s_0)$ and $\log p(s_{t+1}, r_{t+1}|s_t, a_t)$, have no dependence on the parameter $\theta$. Hence, Equation (\ref{eqn:expandPG}) takes shape as
\begin{equation}\label{eqn:expand0_PG}
\nabla_\theta \log \pi_\theta(\tau) = \sum \limits_{t=0}^{T-1} \nabla_\theta \log \pi_\theta(a_t|s_t)~.    
\end{equation}
\noindent In Equation (\ref{eqn:expand0_PG}), $\nabla_\theta \log \pi_\theta(\tau)$ is free from any transition probability terms $p(s',a'|s,a)$, indicating the model-free nature of a policy gradient. 
By substituting the expression (\ref{eqn:expand0_PG}) of $\nabla_\theta \log \pi_\theta(\tau)$ into Equation (\ref{eqn:form0_PG}), we obtain 
\begin{equation}\label{eqn:form1_PG}
    \nabla_\theta J(\theta) = \mathbf{E}_{\tau \sim \pi_\theta} [R(\tau) \sum \limits_{t=0}^{T-1} \nabla_\theta \log \pi_\theta(a_t|s_t)]~. 
\end{equation}
Next, as per the vanilla policy gradient method, i.e. REINFORCE, the return from a trajectory $R(\tau)$ is estimated via the discounted future rewards (rewards-to-go) \cite{REINFORCE1992}. Consider $G_t$ to be the rewards-to-go at the $t^{th}$ transition, defined as: $G_t = \sum_{k=t+1}^{T} \gamma^{k-t-1} r_k$ $\forall t \in \{0,1,...,T-1\}$. By utilizing this reward definition, the policy gradient in Equation (\ref{eqn:form1_PG}) turns into 
\begin{equation}\label{eqn:form2_PGvanilla}
    \nabla_\theta J_{mc}(\theta) = \mathbf{E}_{\tau \sim \pi_\theta} [\sum \limits_{t=0}^{T-1} \nabla_\theta \log \pi_\theta(a_t|s_t) G_t]~. 
\end{equation}
Note that REINFORCE algorithm finds an unbiased estimate of the gradient (\ref{eqn:form2_PGvanilla}) using Monte Carlo sampling, without  assistance of any value function \cite{SuttonPG_nips1999}.

\section{S-REINFORCE Methodology}\label{sec:method}
To achieve tractable policy expressions with an affordable computational budget, the proposed solution procedure involves three main steps: (i) a neural network (NN) is trained exhaustively to update a numerical policy parameters in the gradient direction (\ref{eqn:form2_PGvanilla}), 
(ii) a symbolic regressor (SR) is fitted to extract a symbolic policy at regular intervals along with training the neural network, and
(iii) the fitted symbolic policy is utilized via importance sampling at regular intervals to receive better rewards during training. 
\begin{figure*}[h]
\centering
  \includegraphics[height=3.2in,width=5.8in]{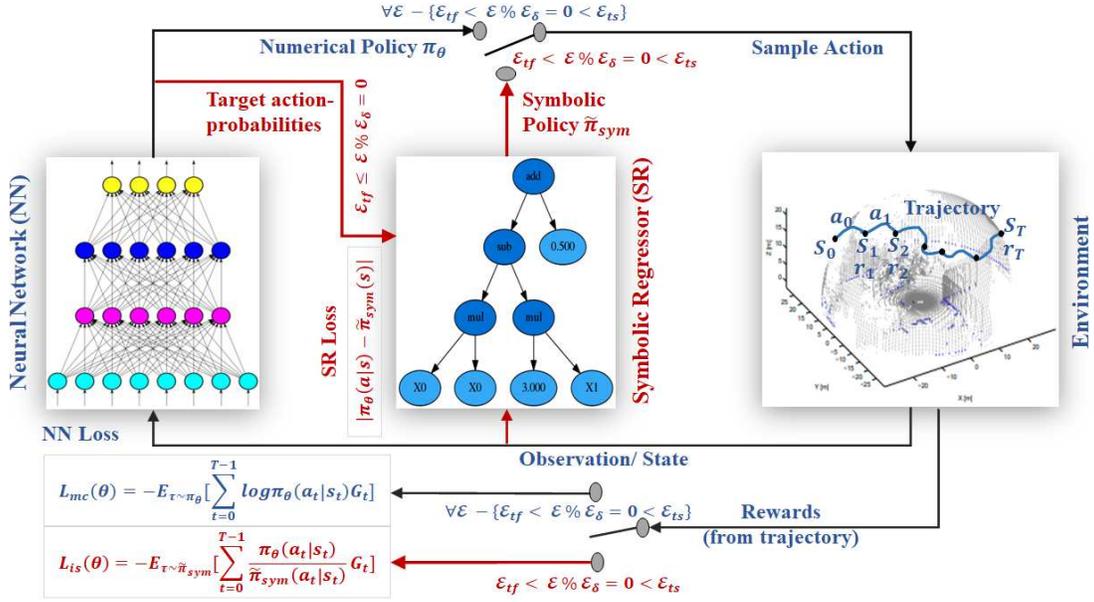}
  \caption{A high-level visualization of the proposed RL-framework composed of two function approximators: NN and SR. Note that the symbol $\%$ in $\mathcal{E}~\%~\mathcal{E}_\delta$ denotes the remainder left over when $\mathcal{E}$ is divided by $\mathcal{E}_\delta$.}
  \label{fig:schematic}
\end{figure*}

\begin{figure*}[h]
\centering
  \includegraphics[height=2.7in,width=6in]{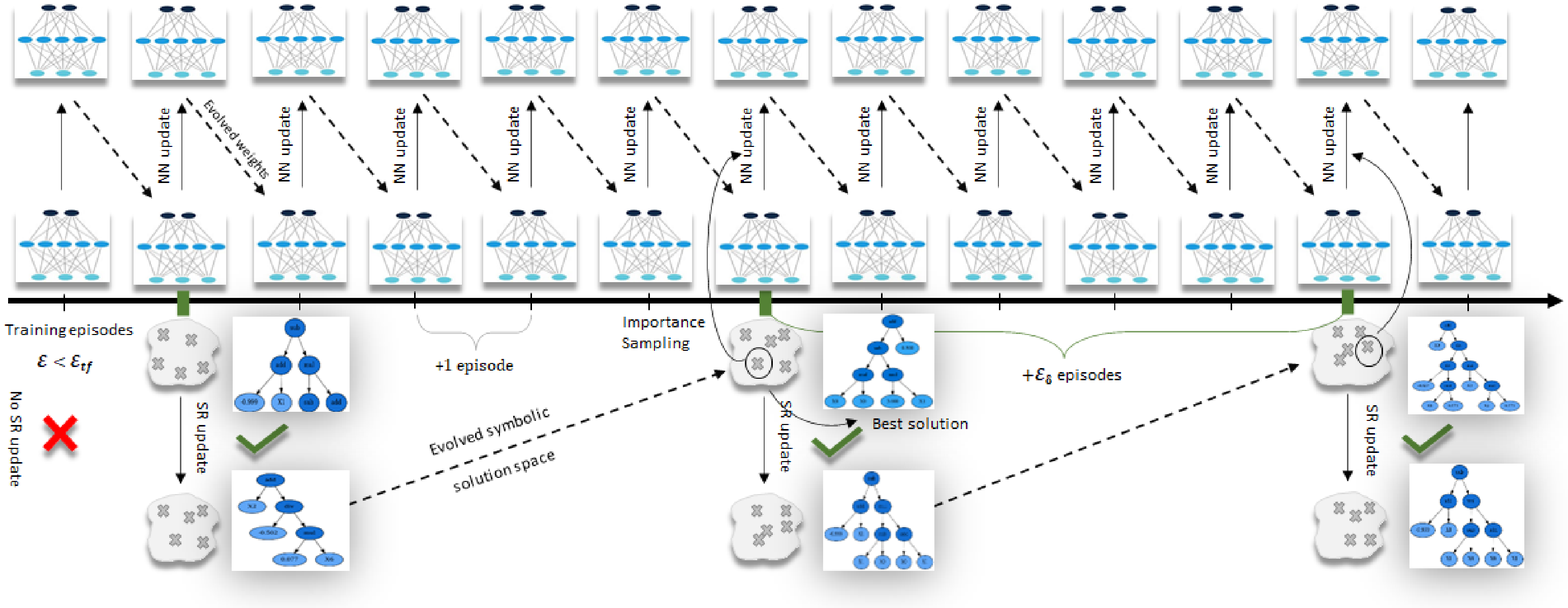}
  \caption{A low-level visualization of the two approximators trained cooperatively. The NN is trained throughout all episodes; the SR fitting starts at episode $\mathcal{E} = \mathcal{E}_{tf}$, and there onwards, this fitting is continued at regular intervals when $\mathcal{E}~\%~ \mathcal{E}_\delta = 0$. Actions are sampled from the numerical policy ($\pi_\theta$) more often except for episodes $\mathcal{E}_{tf} < \mathcal{E}~\%~\mathcal{E}_\delta = 0 < \mathcal{E}_{ts}$, when it is sampled from the symbolic policy ($\tilde{\pi}_{sym}$), and the corresponding NN loss functions are $L_{mc}(\theta)$ and $L_{is}(\theta)$, respectively.}
  \label{fig:timeline}
\end{figure*}

The proposed scheme is depicted in Figure \ref{fig:schematic}, and the details on simultaneous training of the two employed approximators are shown in Figure \ref{fig:timeline}. The pseudocode of the S-REINFORCE algorithm is presented in Algorithm \ref{alg:algo_PGSR}. The roles of different policy approximators and importance sampling are explained in the following. 

\subsection{Approximating Numerical Policy}
It is difficult to find an analytical form of the policy gradient in Equation (\ref{eqn:form2_PGvanilla}). Hence, a function approximator like NN is used to determine a numerical probability distribution (policy) over the possible actions. 
The underlying states are fed to an NN as inputs and it outputs the action probabilities. 
Using Monte Carlo sampling in the training phase, the NN-approximated policy changes by updating weights to minimize the following loss function.
\begin{equation}\label{eqn:lossNN}
    L_{mc}(\theta) = -\mathbf{E}_{\tau \sim \pi_\theta} [\sum \limits_{t=0}^{T-1}  \log \pi_\theta(a_t|s_t) G_t]
\end{equation}
The optimal parameterized policy $\pi_\theta^*$ is determined at the end of the training, when the loss in Equation (\ref{eqn:lossNN}) is minimized. Thus, a trained NN updates the policy along the gradient direction (\ref{eqn:form2_PGvanilla}).  

\subsection{Extracting Symbolic Policy}
The motivation behind using SR lies in its capability of unraveling an interpretable input-output mapping associated with the underlying process \cite{rsos_2022}. For a completed trajectory data, $\{s_t,\pi_\theta(a_t|s_t)\}_{t=0}^{T-1}$, comprised of the input states $s \in \Re^{d}$ and the action-probabilities $\pi_\theta(a|s) \in \Re$, an SR seeks to find the mathematical expression of a policy function: $\tilde{\pi}_{sym}(s): \Re^d \rightarrow \Re$ by minimizing an error functional $e_m(\pi_\theta(a|s), \tilde{\pi}_{sym}(s))$ between the target $\pi_\theta(a|s)$ and the predicted output $\tilde{\pi}_{sym}(s)$ values, where $e_m$ denotes an error metric. The solution to this functional optimization problem is the desired optimal function: $\tilde{\pi}^*_{sym}(s)=arg\min_{\tilde{\pi}_{sym}}e_m(\pi_\theta(a|s), \tilde{\pi}_{sym}(s))$.  

While an NN-based RL agent is being trained, we fit an SR to capture the underlying functional forms of the evolving policy distributions.
Along with training the NN, the SR is started fitting at episode $\mathcal{E}_{tf}$, and thereafter, the fitting is continued at regular intervals of $\mathcal{E}_\delta$ episodes. 
For training episodes $\mathcal{E}_{tf} \leq \mathcal{E}~\%~\mathcal{E}_\delta=0$, the input to the SR contains the states encountered in the preceding trajectory, i.e. $\{ s_t \}_{t=0}^{T-1} \in \Re^{T \times d}$, formed using actions sampled from the NN-approximated policy, and the target is composed of the action-probabilities responsible for the state transitions in that trajectory, i.e. $\{ \pi_\theta(a_t|s_t)\}_{t=0}^{T-1} \in \Re^{T \times 1}$.  

\hl{In practice, genetic programming (GP) is used to solve symbolic regression problems, where a population of candidate programs evolve to attain the optimal solution that best captures the relations between the given input and the targeted output variables.  
Note that each candidate program is represented by a tree made up of numbers, process variables, and symbolic basis functions} \cite{gplearn_2018,rsos_2022}.
\hl{The evolutionary search is carried out with the help of genetic operators: \textit{tournament selection}, \textit{crossover}, and \textit{mutation}.  
The exploitation and exploration of the search space are taken care of by \textit{crossover} and \textit{mutation}, respectively.
Please see the Appendix A for further details on GP.} 

\begin{algorithm}[ht]
\caption{S-REINFORCE: Symbolic Policy Gradient}
\label{alg:algo_PGSR}
\begin{footnotesize}
\begin{algorithmic}
\STATE \textbf{Input}: A differentiable policy parameterization: $\pi_{\theta}(a|s;\theta)$; \textbf{Hyperparameters}: step size $\alpha>0$, episodes $\mathcal{E}_{\max}, \mathcal{E}_{tf}, \mathcal{E}_{\delta}, \mathcal{E}_{ts}$, NN hidden layers \& activations, SR tuning parameters; \textbf{Initialize} numerical policy with random parameters $\theta_0$.\\
\FOR{episodes: $\mathcal{E}=1,2,...,\mathcal{E}_{\max}$}
\STATE $\bullet$ Sample actions from the current policy $\pi_{\theta}$ to generate a trajectory: $\tau = \{s_0, a_0, r_1, s_1,...,s_{T-1}, a_{T-1}, r_T, s_T\}$  of length $T$, containing $\{s_t,a_t,r_{t+1},s_{t+1}\}$  for transitions $t=\{0,...,T-1\}$. 
\STATE $\bullet$ Estimate the Return from trajectory $\tau$ via the discounted future reward $G_t$ at transition $t$: $G_t \leftarrow \sum_{k=t+1}^{T} \gamma^{k-t-1} r_k$ $\forall t \in \{0,1,...,T-1\}$. 
\STATE $\bullet$ Use trajectory $\tau$ to estimate the current policy gradient:\\
\textbf{if} ($\mathcal{E} <= \mathcal{E}_{tf}$) or ($\mathcal{E}_{tf} < \mathcal{E} < \mathcal{E}_{ts}$ and $\mathcal{E} \% \mathcal{E}_\delta \neq 0$) \textbf{then}\\
~~~~Train NN to minimize loss with Monte Carlo sampling: \\ 
$~~~~L_{mc}(\theta)= -J_{mc}(\theta) = - \mathbf{E}_{\tau \sim \pi_\theta}[\sum_{t=0}^{T-1} \log \pi_\theta(a_t|s_t)G_t]$\\
~~~~The associated gradient estimate (\ref{eqn:form2_PGvanilla}), $\hat{g}_{mc}(\theta) \approx \nabla_{\theta}J_{mc}(\theta)$:\\
$~~~~\hat{g}_{mc}(\theta) \leftarrow \mathbf{E}_{\tau \sim \pi_\theta}[\sum_{t=0}^{T-1} \nabla_{\theta}\log \pi_\theta(a_t|s_t)G_t]$\\
\textbf{else}\\
~~~~Train NN to minimize loss with importance sampling: \\
$~~~~L_{is}(\theta)=  -J_{is}(\theta)= -\mathbf{E}_{\tau \sim \tilde{\pi}_{sym}}[\sum_{t=0}^{T-1}\frac{\pi_{\theta}(a_t|s_t)}{\tilde{\pi}_{sym}(a_t|s_t)}G_t]$\\
~~~~The associated gradient estimate (\ref{eqn:estPG_is}), $\hat{g}_{is}(\theta) \approx \nabla_{\theta}J_{is}(\theta)$:\\
$~~~~\hat{g}_{is}(\theta) \leftarrow  \mathbf{E}_{\tau \sim \tilde{\pi}_{sym}}[\sum_{t=0}^{T-1} \frac{\pi_{\theta}(a_t|s_t)}{\tilde{\pi}_{sym}(a_t|s_t)}\nabla_{\theta}\log \pi_\theta(a_t|s_t)G_t$]\\
\textbf{end if}\\
\STATE $\bullet$ Update the weights $\theta$ of the current policy:\\ 
\textbf{if} ($\mathcal{E}_{tf} < \mathcal{E} < \mathcal{E}_{ts}$) and ($\mathcal{E} \% \mathcal{E}_\delta = 0$) \textbf{then}\\
$~~~~\theta \leftarrow \theta + \alpha \hat{g}_{is}(\theta)$ \\
\textbf{else}\\
$~~~~\theta \leftarrow \theta + \alpha \hat{g}_{mc}(\theta)$ \\
\textbf{end if}\\
\STATE $\bullet$ Transfer knowledge from NN to SR: \\
\textbf{if} ($\mathcal{E} >= \mathcal{E}_{tf}$ and $\mathcal{E} \% \mathcal{E}_\delta = 0$) \textbf{then}\\
~~~Train SR to obtain the functional policy: \\
~~~\textit{Input:} $S_\mathcal{E}=\{s_0,...,s_T\}$\\
~~~\textit{Target:} $P_\mathcal{E}=\{\pi_{\theta}(a_0|s_0),...,\pi_{\theta}(a_{T-1}|s_{T-1})\}$\\
$~~~\tilde{\pi}_{sym} \leftarrow SR(S_\mathcal{E}, P_\mathcal{E})$\\
\textbf{end if}\\
\ENDFOR
\STATE \textbf{Return}: Learnt numeric policy $\pi_{\theta}$ and symbolic policy $\tilde{\pi}_{sym}$
\end{algorithmic}
\end{footnotesize}
\end{algorithm}

\subsection{Increasing Sampling Efficiency}
The policy gradient methods learn an optimal policy by updating the policy parameters along the gradient of the expected return, given by: 
$\nabla_\theta J(\pi_\theta) = \mathbf{E}_{s \sim \rho^\pi, a \sim \pi_\theta} [\sum_{t=0}^{T-1} \nabla_\theta \log \pi_\theta(a_t|s_t) G_t]$. 
Consider that data $\mathcal{D}$ contains the set of state-action pairs encountered in a trajectory sampled using the current policy \cite{aamas_2019}.
Then, the policy gradient estimator takes shape as:
\begin{equation*}
\nabla_\theta J_{mc}(\mathcal{D}) = \mathbf{E}_{s \sim \rho^\mathcal{D}, a \sim \pi_\mathcal{D}} [\sum_{t=0}^{T-1} \nabla_\theta \log \pi_\theta(a_t|s_t) G_t]~,
\end{equation*}
where $\nabla_\theta J_{mc}$ is the Monte Carlo estimator of the policy gradient. $\nabla_\theta J_{mc}$ is an unbiased estimator of $\nabla_\theta J(\pi_\theta)$, indicating that the gradient estimate would be accurate for repeatedly sampled batches of data. However, a single batch of collected data considers only a limited number of states visited prior to evaluating the gradient, which leads to inaccurate estimates. 
Earlier research \cite{aamas_2019} realized that the sampling error arises because the expectation in $\nabla_\theta J_{mc}$ is taken over a deviated action distribution $\pi_\mathcal{D}$ rather than the actual action distribution $\pi_\theta$.  
The correct state distribution is usually unknown, however, the availability of the correct action distribution ($\pi_\theta$) can be leveraged via importance sampling to design a sampling correction \cite{aamas_2019}. 
The gradient estimate's variance due to the stochastic nature of action selection is also reduced by such a sampling correction \cite{aamas_2019}.    

In other words, for a trajectory containing finite state-transitions, the samples collected using only the current policy are not sufficient for an accurate gradient estimation. 
Instead, a policy mapping $\tilde{\pi}_{sym}$ fitted with the preceding trajectory data can be used to improve the estimation accuracy.
When actions are sampled from $\tilde{\pi}_{sym}$, which is close to $\pi_\mathcal{D}$, then this sampling correction eliminates variance in the action selection \cite{aamas_2019}. A detailed variance analysis is available in the Appendix B.
In this regard, the expected reward of a trajectory is computed via importance sampling (IS) that utilizes a different policy for sample collection, which is compensated by a ratio of probability distributions in the objective as follows.
\begin{equation}\label{eqn:objIS_ExpRt}
    J(\theta) = \mathbf{E}_{\tau \sim \tilde{\pi}_{sym}} [\frac{\pi_\theta(\tau)}{\tilde{\pi}_{sym}(\tau)}R(\tau)]~ 
\end{equation}
The associated policy gradient is given by 
\begin{eqnarray} \nonumber
& \nabla_\theta J(\theta)= \mathbf{E}_{\tau \sim \tilde{\pi}_{sym}} [\frac{\nabla_\theta \pi_\theta(\tau)}{\tilde{\pi}_{sym}(\tau)}R(\tau)] \\ \label{eqn:PGrad0_IStraj}
&~or,~ \nabla_\theta J(\theta) = \mathbf{E}_{\tau \sim \tilde{\pi}_{sym}} [\frac{\pi_\theta(\tau)}{\tilde{\pi}_{sym}(\tau)} \nabla_\theta \log \pi_\theta(\tau) R(\tau)],
\end{eqnarray}
where the ratio $\frac{\pi_\theta(\tau)}{\tilde{\pi}_{sym}(\tau)}= \frac{\prod_{i=0}^{T-1} \pi_\theta(a_t|s_t)}{\prod_{i=0}^{T-1} \tilde{\pi}_{sym}(a_t|s_t)}$
can further be simplified by means of the likelihood ratios, using Markov's principle of causality \cite{nips_2010}. Consequently, the policy gradient form in Equation (\ref{eqn:form2_PGvanilla}) turns into  
\begin{equation}\label{eqn:estPG_is}
    \nabla_\theta J_{is}(\theta) = \mathbf{E}_{\tau \sim \tilde{\pi}_{sym}} [\sum \limits_{t=0}^{T-1} \frac{\pi_\theta(a_t|s_t)}{\tilde{\pi}_{sym}(a_t|s_t)}\nabla_\theta \log \pi_\theta(a_t|s_t) G_t].
\end{equation}
Let us now analyze the variance induced by the IS-based estimator: $Var[\frac{\pi_\theta(\tau)}{\tilde{\pi}_{sym}(\tau)}R(\tau)]$ 
\begin{eqnarray}\nonumber
& = \mathbf{E}_{\tau \sim \tilde{\pi}_{sym}}[\frac{\pi^2_\theta(\tau)}{\tilde{\pi}^2_{sym}(\tau)}R^2(\tau)] - \mathbf{E}^2_{\tau \sim \tilde{\pi}_{sym}}[\frac{\pi_\theta(\tau)}{\tilde{\pi}_{sym}(\tau)}R(\tau)]^2 \\ \label{eqn:varIS}
& or, Var[\frac{\pi_\theta(\tau)}{\tilde{\pi}_{sym}(\tau)}R(\tau)] = \mathbf{E}_{\tau \sim \pi_\theta}[\frac{\pi_\theta(\tau)}{\tilde{\pi}_{sym}(\tau)}R^2(\tau)] - J^2 .
\end{eqnarray}
Equation (\ref{eqn:varIS}) reveals that the IS-induced variance becomes high when the distributions $\pi_\theta$ and $\tilde{\pi}_{sym}$ are widely different. According to the earlier research \cite{nips_2010}, \cite{aamas_2019}, the variance is minimum when $\tilde{\pi}_{sym}(\tau) \propto \pi_\theta(\tau)|R(\tau)|$ is satisfied \cite{impSample_review2010}, which indicates that IS should be applied when an RL agent receives good rewards during the learning process.

\section{Empirical Results}\label{sec:res}
In the following, we present the results obtained using the proposed methodology applied onto different dynamic decision-making problems with low to high-dimensional action spaces.  

\begin{table*}[ht]
\scriptsize
\centering
\begin{tabular}{cccccc}
\hline
Scenario & NN parameters & SR parameters & SR occurrence & Importance sampling  &  \\ \hline \hline
  & (FC-128 with tanh)+  & population\_size$=2000$, tournament\_size$=20$,  &  &  &     \\
Cartpole  & (FC-128 with softmax) & p\_crossover $=0.7$, p\_subtree\_mutation$=0.1$, & $\mathcal{E}=400$ &  $500 \leq \mathcal{E} \leq 1800$    &   \\
  & learning rate$=3\times 10^{-4}$ &  p\_hoist\_mutation$=0.05$, p\_point\_mutation$=0.1$  & $\mathcal{E} > 400$ and $\mathcal{E}~\%~10=0$ & and $\mathcal{E}~\%~10=0$ &     \\
  &   & Basis: [add, sub, mul, div, inv, cos] & & &  \\\hline

  & (FC-128 with tanh)+  & population\_size$=2000$, tournament\_size$=50$,  &  &  &     \\
Acrobot & (FC-128 with tanh)+ &  p\_crossover $=0.7$, p\_subtree\_mutation$=0.1$,  &  $\mathcal{E}=400$ &  $500 \leq \mathcal{E} \leq 1800$    &   \\
  & (FC-128 with softmax) &  p\_hoist\_mutation$=0.05$, p\_point\_mutation$=0.1$ & $\mathcal{E} > 400$ and $\mathcal{E}~\%~10=0$  &  and $\mathcal{E}~\%~10=0$  & \\ 
    & learning rate$=5\times 10^{-4}$ &  Basis: [add, sub, mul, div, inv, cos]   &   &   &   \\ \hline

  & (FC-128 with tanh)+  & population\_size$=2000$, tournament\_size$=20$,  &  &  &     \\
Lunar Lander & (FC-128 with tanh)+ &  p\_crossover $=0.7$, p\_subtree\_mutation$=0.1$,  &  $\mathcal{E}=400$ &  $500 \leq \mathcal{E} \leq 1800$    &   \\
  & FC-128$^\#$ &  p\_hoist\_mutation$=0.05$, p\_point\_mutation$=0.1$ & $\mathcal{E} > 400$ and $\mathcal{E}~\%~10=0$  &  and $\mathcal{E}~\%~10=0$  & \\ 
    & learning rate$=5\times 10^{-4}$ &  Basis: [add, sub, mul, div, min, max]   &   &   &   \\ 
\hline
\end{tabular}
\caption{Implementation details across three different environments. Here, `FC-128' denotes a fully-connected layer with 128 neurons. Note: $^\#$ The output layers are different for mean and variance approximations contributing to a Gaussian action-probability distribution.}
\label{tab:implement}
\end{table*}

\noindent  \textbf{Cartpole:}  In the Cartpole-v0 environment, a pole is attached to a cart moving along a frictionless track. The pole starts upright and the RL agent aims to prevent it from falling over by applying a force of $-1$ or $+1$ to the cart. 
The environment consists of four states: cart position and velocity, pole angle and angular velocity, and the action space is discrete with two possibilities: left and right.
The agent is rewarded with $(+1)$ for every time step the pole remains upright.
\begin{figure}[h]
  \includegraphics[height=2.4in,width=3.65in]{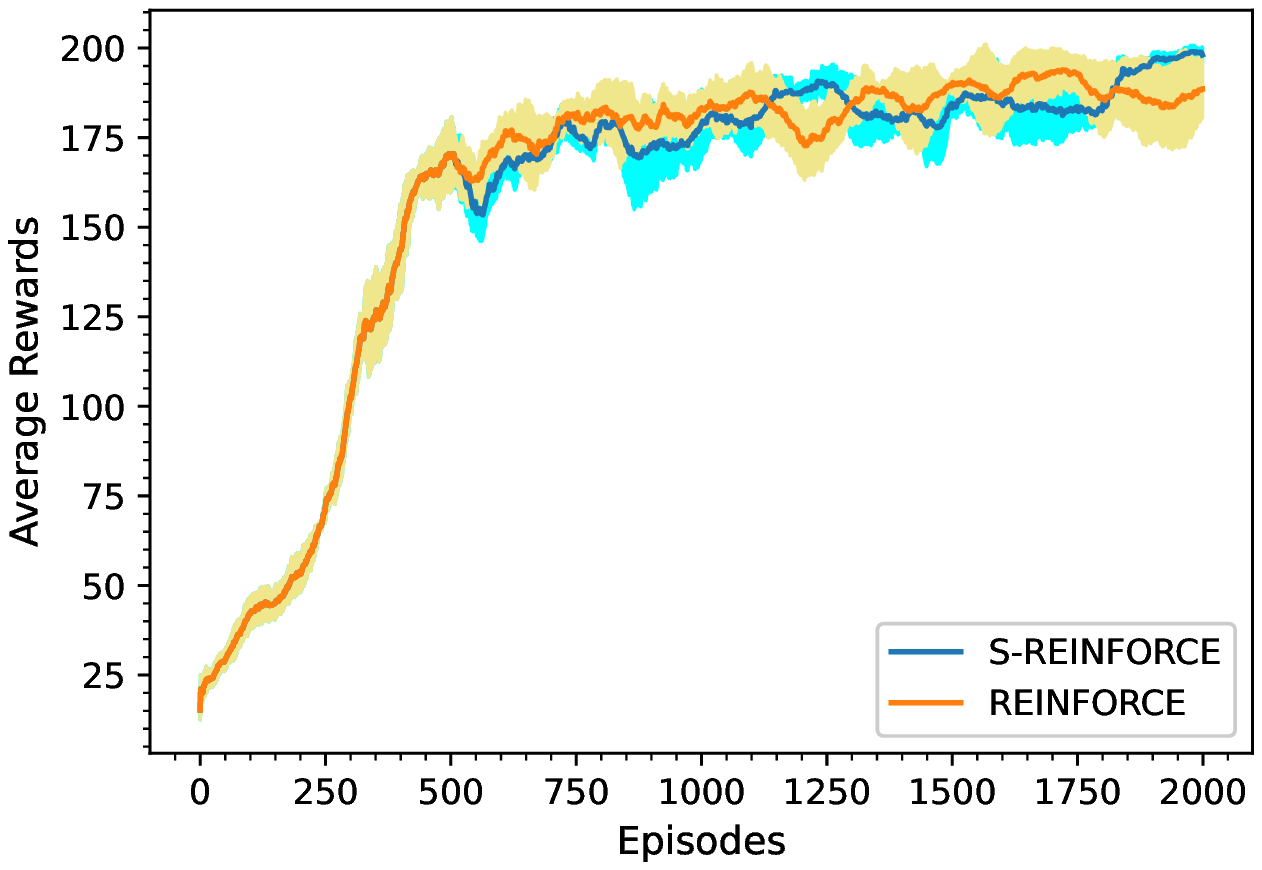}
  \caption{Performance comparison of the rewards received by different RL agents in the Cartpole environment. The shaded region depicts the mean $\pm$ the standard deviation of the moving average (of size $50$) rewards across five different random seeds.}
  \label{fig:CP}
\end{figure}


\hl{In this case, an SR is fitted along with training an NN and the related implementation details are shown in Table} \ref{tab:implement}.
\hl{The reward profiles obtained with REINFORCE and S-REINFORCE algorithms are shown in Figure} \ref{fig:CP}, \hl{wherein an improved reward profile is observed in the case of S-REINFORCE.}  
Figure \ref{fig:CP} exhibits that the rewards achieved with REINFORCE falls slightly after $1700$ episodes, which is alleviated by S-REINFORCE.
After $2000$ episodes of training, the symbolically approximated policies, $\pi(a_0)$ and $\pi(a_1)$, are shown in Table \ref{tab:result}.
Note that the action-probabilities are restricted to $0.5$ in case they are negative for some permissible states.
\hl{These functional forms of the action-probability distributions reveal that applying a negative or a positive force mainly depends on the two states, i.e. pole angle and pole angular velocity. Also, a noticeable amount of offset is present in the probability expressions.}    

\noindent \textbf{Acrobot:}  Acrobot-v1 environment simulates a chain-like system made up of two linearly connected links with one end fixed. The joint between the two links is actuated, such that the free end of the linear chain swings above a desired height while starting from a downward hanging position. The state space is $6$-dimensional, which provides information about the two rotational joint angles and their angular velocities, and it is bounded by: $[-1, -1, -1, -1, -12.57, -28.27]$ to $[1, 1, 1, 1, 12.57, 28.27]$.
The action space is discrete with three possibilities representing the torques applied on the actuated joint, as follows: \textit{action 0:} apply $-1$ torque, \textit{action 1:} apply $0$ torque, and \textit{action 2:} apply $+1$ torque.  
The RL agent is rewarded with $0$ when the free end reaches the desired height, and it is penalized with $-1$ at every step if the targeted height is not attained.  
\begin{figure}[h]
\centering
  \includegraphics[height=2.4in,width=3.65in]{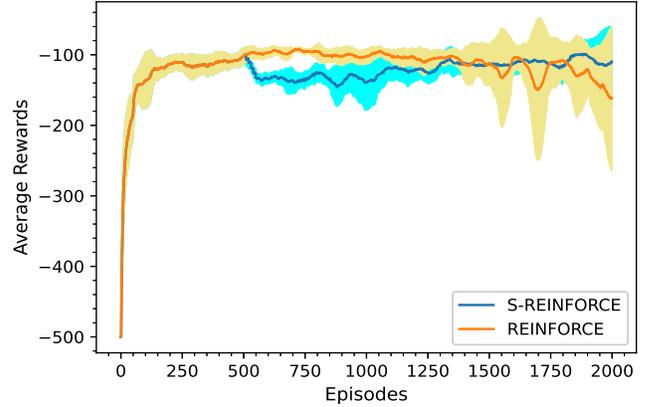}
  \caption{Performance comparison of the rewards received by  different RL agents in the Acrobot environment.} 
  \label{fig:AB}
\end{figure}


\hl{In this case, two identical SRs are fitted along with training an NN and the related implementation details are shown in Table} \ref{tab:implement}.
\hl{The vanilla policy gradient agent takes thousands of training episodes to solve the Acrobot environment} \cite{benchmark2016,Green2018}. 
\hl{Nevertheless, the reward profiles, obtained with REINFORCE and S-REINFORCE algorithms, up to $2000$ episodes are shown in Figure} \ref{fig:AB}. 
In Figure \ref{fig:AB}, \hl{the performance of the REINFORCE agent deteriorates after around $1500$ episodes}, which is gently recovered by the S-REINFORCE agent.
After $2000$ episodes of training, the symbolically approximated policies, $\pi(a_0),~\pi(a_1)$ and $\pi(a_2)$, are shown in Table \ref{tab:result}.
\hl{The action-probability expressions indicate that applying a negative torque is more probable than applying a positive torque, and a zero torque is the least probable action. Also, such an action selection is only influenced by the fifth state.}   


\noindent  \textbf{Lunar Lander:} LunarLanderContinuous-v2 environment simulates a robotic control task, where a rover aims to land on the landing pad between two flags. 
The underlying state space is 8-dimensional, which is comprised of horizontal and vertical positions and velocities, angular position and velocity, and binary numbers indicating whether any of the rover legs are in contact with the ground. This state space is real-valued and bounded within $(-\infty, \infty)$.   
The action space is 2-dimensional. The first action corresponds to the throttle of the main engine and it varies continuously between $-1$ and $+1$; the main engine is turned off when this action is below $0$ and the throttle scales affinely from $50\%$ to $100\%$ for actions in the range $(0,1]$. 
The second action corresponds to the throttle of the lateral engine; the left booster is fired when it is below $-0.5$ and the right booster is fired when it is above $0.5$, and the throttle scales affinely from $50\%$ to $100\%$ for actions in ranges $[-1, -0.5]$ and $[0.5, 1]$.
\begin{figure}[h]
\centering
  \includegraphics[height=2.4in,width=3.65in]{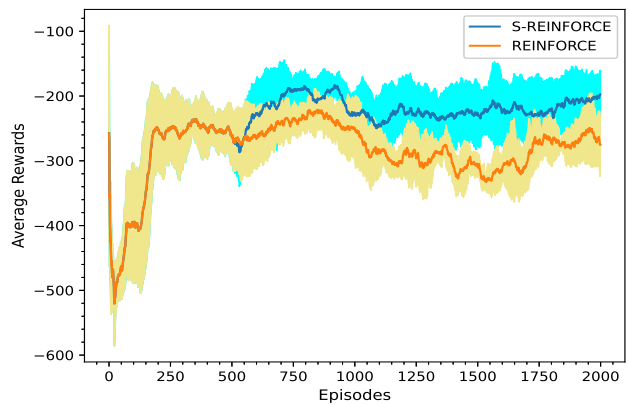}
  \caption{Performance comparison of the rewards received by different RL agents in the Lunar Lander environment.}
  \label{fig:rewardsLL}
\end{figure}



\hl{In this case, along with training two NNs, four identical SRs are fitted to find the symbolic mappings of the means and the variances defining two Gaussian action-probability distributions. The related implementation details are highlighted in Table} \ref{tab:implement}.
\hl{It takes around 20000 episodes to solve the LunarLander environment with the vanilla policy gradient algorithm, i.e. REINFORCE} \cite{stat2017}.
\hl{However, in this study, we present a performance comparison between REINFORCE and S-REINFORCE agents up to 2000 episodes.} 
In Figure \ref{fig:rewardsLL}, the rewards obtained with REINFORCE increase gradually after around $500$ episodes, however, this increase is not sustained well afterwards. 
On the other hand, the increase in the rewards is well maintained by S-REINFORCE and the reward profile has less fluctuations than that of REINFORCE.      
After $2000$ episodes of training, the symbolic policies are shown in Table \ref{tab:result}.
\hl{These mean and variance expressions indicate that the expected first action depends on five states, $s_0, s_1, s_3, s_4, s_5$, though the expected second action depends on four states, $s_1, s_4, s_5, s_7$. Also, the choice of the actions does not depend on $s_2$.}

\begin{table*}
\footnotesize
\centering
\begin{tabular}{c|c|c|c} \hline
Scenario &  Reward$_p$  &  Reward$_b$ &  Policy Expression \\ \hline \hline
 & & & \\
Cartpole & $198.096$ & $188.596$ & $\pi(a_0)=0.52-2s_2-0.595s_3$\\
&  $\pm 2.174$  &  $\pm 8.080$ & $\pi(a_1)=0.48 + 2s_2 + 0.595s_3$ \\ 
& & & \\ \hline 
& & & \\
& & & $\pi(a_0)=0.404 - 0.155s_5-0.31cos(s_5-1.976)$\\
Acrobot & $-109.852$ & $-161.512$ & $\pi(a_1)=0.006$ \\
& $\pm 45.951$ & $\pm 101.985$ & $\pi(a_2)=0.59+0.155s_5+0.31cos(s_5-1.976)$ \\ 
& & & \\ \hline
& & & \\
& & & $\mu(a_0)=3.61-s_0-7.04s_1-5s_3-2s_4-1.45s_5-s_5^2$\\
LunarLander & $-274.914$ & $-294.721$ & $\sigma(a_0)= max(0.01, 0.95 \frac{s_4 s_5 s_7}{s_1})$\\ 
& $\pm 83.351$ & $\pm 66.456$ & $\mu(a_1)=0.95 \frac{s_4 s_5 s_7}{s_1}$ \\
& & & $\sigma(a_1)=max(0.01, 0.95 \frac{s_4 s_5 s_7}{s_1})$ \\
& & & \\ \hline
\end{tabular}
\caption{Performance evaluation of S-REINFORCE in different scenarios: Reward$_p$ and Reward$_b$ denote the rewards achieved with the proposed S-REINFORCE and the baseline REINFORCE algorithms, respectively, after 2000 episodes of training across five different random seeds, which are described in the form of (mean $\pm$ standard deviation). In case of a continuous action space, $\mu$ and $\sigma$ denote the mean and the standard deviation of a Gaussian action-probability distribution, respectively. For each scenario, the reported policy expression represents the best performing symbolic policy among five different seeds.}
\label{tab:result}
\end{table*}

\begin{figure}[h]
  \includegraphics[height=2.35in,width=3.54in]{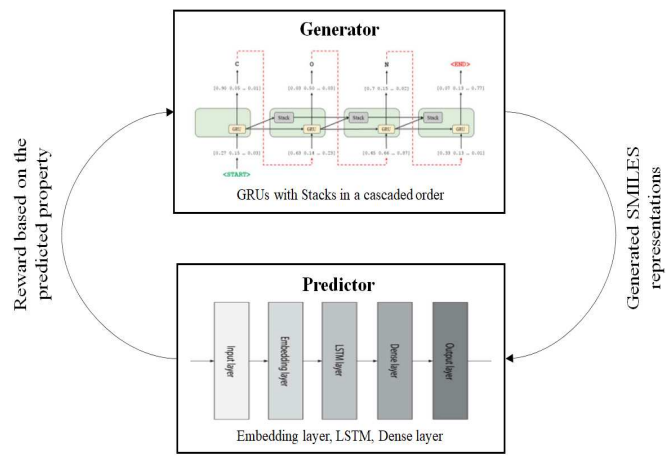}
  \caption{\hl{Flow of optimal generation: a pre-trained generator is fine-tuned using RL to produce molecules, represented by SMILES, with the desired property: $0 < logP < 5$.}}
  \label{fig:release_model}
\end{figure}

\noindent  \textbf{Reinforcement Learning for Structural Evolution (ReLeaSE):} ReLeaSE environment consists of two deep neural networks, a \textit{generator} and a \textit{predictor}. In the pre-training phase, the generator learns to produce chemically viable molecules and the predictor learns to evaluate the generator’s performance using a supervised learning algorithm \cite{denovoRL_Science2018}. Next, an RL-agent is leveraged to attain a desired property by fine-tuning the pre-trained generator, as shown in Figure \ref{fig:release_model}. 

The SMILES representation of a molecule here denotes the state. Starting from an empty string (initial state), the \textit{generator} takes a sequence of appropriate actions to construct a reasonable SMILES string (final state). An action is represented by an addition of a token from a set of alphabets, numbers and special characters, which is sampled from the policy, i.e. a probability distribution over all the possible actions. These probabilities are produced by a stack-augmented recurrent neural network (Stack-RNN) architecture with memory augmentation, which helps it to learn long-range dependencies in the generated SMILES strings with reference to chemically feasible structures from a DeepChem-based decoder. The \textit{predictor} is used to estimate physical, chemical or biological properties of the generated molecules. Here, the targeted property is the “logP” value of a molecule, which signifies the hydrophobicity of a compound. In this case study, the goal is to generate orally bioavailable molecules/compounds that have $logP < 5$ as per the rule of Lipinski \cite{ReLeaSE_Github}. Given an input SMILES string representing a specific molecule, the predictor outputs an estimate of its $logP$ value using an embedding layer, an LSTM layer and two dense layers \cite{denovoRL_Science2018}. 

Once pre-training is over, an RL agent learns to bias the generator incrementally by maximizing the rewards given to it, such that the predicted logP values of the generated molecules align with the reference $logP$ profile \cite{denovoRL_Science2018}. 
The reward assignment is adopted from the earlier study where ReLeaSE was proposed \cite{denovoRL_Science2018}, \cite{ReLeaSE_Github}, with the reward design:
$R=11$ if $1 < \text{logP} < 4$ and $R=1$ otherwise. 
An NN was employed to numerically approximate the underlying action-probability distribution that updates iteratively according to the REINFORCE policy gradient method. However, the lack of interpretability of a numerical policy representation motivates us to call for a symbolic regression along with training NN. To fit an SR, the Morgan fingerprints \cite{fpMorgan_2010} of all the molecules formed in a trajectory are fed as inputs and the probabilities of sampling tokens for all molecule formations in that trajectory are set to the target values. 
In the current S-REINFORCE framework, the state space contains $1024$ bits Morgan fingerprints where each element is a binary number, and the action space is $45$-dimensional with a variety of choices for different token additions.  

\noindent  \textit{Implementation:}  The \textit{generator} and \textit{predictor} architectures are adopted from the earlier study \cite{denovoRL_Science2018}, \cite{ReLeaSE_Github}.
The parameters associated with RL are: 
number of iterations (outer loops), i.e. n\_iterations $= 50$, 
number of trajectories/molecules formed with the \textit{generator} policies, i.e. n\_policy $= 15$,
number of batches under every generated policy, i.e. n\_policy\_replay $= 10$, and
number of molecules generated for $\log P$ property prediction, i.e. n\_to\_generate $= 1000$.
It is worth noting that the training loss is averaged over all the batches, so the molecule properties are monitored over total $60 \times 15=900$ training episodes.  
Here, $45$ distinct SRs are employed to capture functional relations in all the action-probabilities.   
The parameters associated with each SR are: population\_size$=500$, tournament\_size=20, p\_crossover $=0.7$, p\_subtree\_mutation$=0.1$, p\_hoist\_mutation$=0.05$, p\_point\_mutation$=0.1$, parsimony\_coefficient$=0.005$.
The basis function set used in the symbolic regression is: [add, sub, mul, div, cos].
The SRs are fitted for: (n\_iterations $>= 5$ and $\% 5=0$) and (n\_policy\_replay $\% 10=0$), 
and importance sampling is applied for: (n\_iterations $> 10$ and $\% 5=0$) and (n\_policy\_replay $\% 10=0$). 

\begin{figure}[h]
\centering
\begin{subfigure}{0.5\textwidth}
  \includegraphics[height=2.35in,width=3.65in]{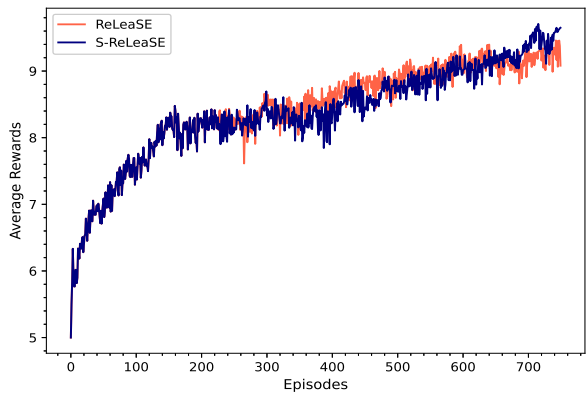}
  \caption{Rewards received in ReLeaSE environment.}
  \label{fig:symRelease}
\end{subfigure}
\hfill
\begin{subfigure}{0.5\textwidth}
  \includegraphics[height=2.35in,width=3.65in]{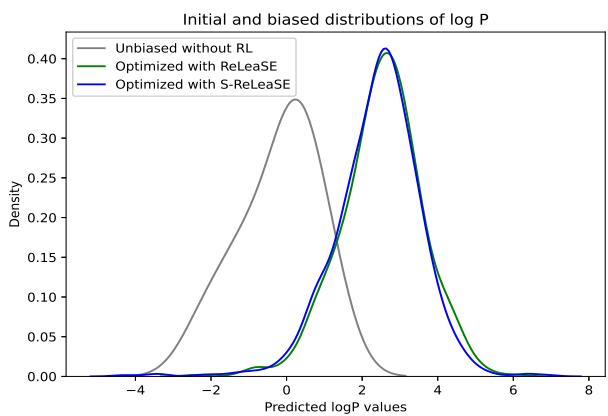}
  \caption{\hl{Predicted properties with and without RL.}}
  \label{fig:predS-ReLeaSE}
\end{subfigure}
\caption{Performance of RL in molecule structural evolution.}
\label{fig:S-ReLeaSE}
\end{figure}

\noindent \textit{Outcome:} After $750$ episodes of training, the maximum average reward ($9.5$) achieved with the symbolic ReLeaSE (S-ReLeaSE) is slightly better than that ($9.2$) of the original ReLeaSE, although the slopes of both the reward profiles are similar, as shown in Figure \ref{fig:symRelease}.   
Here, importance sampling does not give much improvement in rewards as numerous visited states from multiple trajectories are considered in repeatedly sampled batches of data prior to updating the policies.  
Figure \ref{fig:predS-ReLeaSE} exhibits that the logP distribution over $1000$ molecules predicted with the biased (fine-tuned) generator shifts into the desired zone effectively, as compared to the logP distribution predicted with the unbiased (pre-trained) generator.  
\hl{However, the molecular logP distributions predicted with ReLeaSE and S-ReLeaSE are almost identical.} 
\hl{The areas under the density curves indicate that the generator fine-tuned with RL is able to produce much more desired molecules than the unbiased generator. Four sample molecules produced by the fine-tuned generator are: SMILES $\rightarrow$
``Brc1ccccc1, C=C(C=CC)Sc1ccccc1, C=CCN(C)CCCOc1ccccc1, CC(=O)CN1c2ccccc2-c2ccccc2C1c1ccccc1" with their logP values as: 2.45, 3.87, 2.57, 4.85, respectively.}                    

The policy expressions achieved with S-ReLeaSE for the first three action-probabilities, are as follows.  
\begin{footnotesize}
\begin{align*}
& \log \pi(a_0) = \frac{cos(s_{577})}{cos^3(s_{287})-cos(s_{234})} \times \{cos^3( s_{572} \times s_{342} + \\ & cos^2(cos(s_{577})+cos(s_{583} s_{33}))-s_{248})+cos(s_{697})\};\\
& \log \pi(a_1) = s_{697}-cos(s_{408}) - cos(s_{621}) - s_{936} - s_{131} - cos(s_{408});\\
& \log \pi(a_2) = \{cos(s_{569})+cos(s_{604})\}\times cos^{-3}(s_{604})/ \\ & \{cos^2(s_{569}) - cos( (s_{757}-cos(s_{822})) \times (cos^2(s_{154})-cos(s_{848})) )\}.
\end{align*}
\end{footnotesize}
The above functional policies are normalized prior to sampling actions from them, such to maintain $\sum_{i=0}^{44} \pi(a_i)=1$.
\hl{These action-probabilities, $\pi(a_0), \pi(a_1),\pi(a_2)$, are uncorrelated as they do not depend on any common states. 
Interestingly, each policy relates to a limited number of states.} 

\noindent  \textbf{Discussion:}  The achieved results justify that the proposed S-REINFORCE algorithm can produce explicit policy expressions with causal relationships, and the obtained average reward profiles contain less fluctuations than the regular REINFORCE. 
Also, after completion of training in different environments, the final rewards received by the proposed RL agent is higher than the baseline RL agent, as shown in Table \ref{tab:result}. 
Rather than using off-policy data \cite{offPolicyIS_2013}, importance sampling is applied in this study to correct the policy shift using on-policy data.
Figures \ref{fig:AB} and \ref{fig:rewardsLL} reveal noticeable dips in the reward profiles after $500$ episodes due to inadequate fitting of the associated SRs.
As the accuracy of the symbolic regression improves, the adopted importance sampling becomes more effective, resulting in a faster learning. 
Therefore, achieving optimal performance requires accurate symbolic regression using on-policy data in conjunction with training NN. 
The choice of basis functions and hyper-parameters in SR is crucial for achieving the optimal performance, in addition to the proper tuning of knowledge transfer-related parameters, such as: $\mathcal{E}_{tf}$, $\mathcal{E}_\delta$ and $\mathcal{E}_{ts}$.
Future research should focus on rigorously exploring the tunable parameters to improve the S-REINFORCE agent's performance in complex real-world environments.

\section{Conclusion}\label{sec:conclude}
The proposed algorithm is a breakthrough in the field of sequential decision-making tasks, as it not only improves the rewards received by an RL agent but also generates interpretable policy expressions. This is achieved by simultaneously training both numerical and symbolic approximators, which enables the algorithm to handle both low and high dimensional action spaces. Our methodology has been tested in various dynamic scenarios, and has been shown to be effective in generating appropriate control policies. This work serves as a proof-of-concept for combining numerical and symbolic policy approximators into an RL framework. While this is a significant step forward, we recognize that there is still much work to be done. In the future, we plan to extend the underlying idea to more advanced RL techniques that require both value and policy approximations. 


\section*{Acknowledgements}
This study is supported by the Accelerated Materials Development for Manufacturing Program at A*STAR via the AME Programmatic Fund by the Agency for Science, Technology and Research under Grant No. A1898b0043.

\appendix

\section{Genetic Programming (GP)}
\hl{The candidate programs in GP evolve over generations until a prescribed stopping criteria is met.
The performance of GP is sensitive to the associated hyper-parameters. 
In this context, to achieve an efficient outcome without involving high computational overhead, the total number of programs in each generation and the number of programs taking part in a tournament to go to the next generation are the basic factors.
For search space exploitation, a key parameter, p\_crossover, represents the probability of performing crossover to transfer valuable subtrees from the parents (tournament winners) to the offspring in the next generation. 
For search space exploration, there are three kinds of mutation parameters: (i)  p\_subtree\_mutation is the probability of replacing a significant amount of genetic material from the tournament winners to introduce extinct functions into a population of candidate solutions, so it is an aggressive mutation parameter; 
(ii) p\_hoist\_mutation denotes the probability of removing genetic material by hoisting a random subtree into the original subtree's location, thus it is a bloat-resisting mutation parameter; and (iii) p\_point\_mutation  denotes the probability of replacing random nodes from the tournament winners to introduce new functions.
Besides, the constant, parsimony\_coefficient, penalizes large programs by adjusting their fitness to be less favorable for the selection process.}

\hl{The choice of basis functions or elementary compositions plays a pivotal role in practice. Several trial and error simulations are conducted to decide on the appropriate library of basis functions. Depending on the complexity of the problem and the underlying environment, this library can be adjusted (reduced or enhanced). 
Note that some of the basis functions like 'sqrt', 'inv', 'exp' are protected to avoid numerical overflow errors.}

\section{Variance Analysis}
Here, we provide the variance analysis of the currently adopted gradient estimators. 
Consider $\mathbb{S}$ and $\mathbb{A}$ to be the random variables associated with the states and actions, respectively. Next, consider a generic form of the policy gradient as: $g(\{\mathbb{S}, \mathbb{A}\})=$  
\begin{equation*}\label{eqn:gradEst_generic}
     =  \sum_{s \in \mathcal{S}} \rho^{\mathcal{D}}(s) \sum_{a \in \mathcal{A}} \pi_\mathcal{D}(a|s) w(s,a) \hat{G}(s,a) \nabla_\theta \log \pi_\theta(a|s),
\end{equation*}
where $w(s,a)=1$ for the Monte Carlo estimator $g_{mc}$ and $w(s,a)=\frac{\pi_\theta(a|s)}{\tilde{\pi}_{sym}(a|s)}$ for the importance sampling based estimator $g_{is}$; $\hat{G}(s,a)$ is the estimated return.   
The variance of the estimated gradient, $g(\{\mathbb{S}, \mathbb{A}\})$, can be decomposed using the total law of variance \cite{aamas_2019}, as follows. 
\begin{eqnarray*}\label{eqn:VarDecomp}
&    \mathbf{Var}_{\mathbb{S}, \mathbb{A}} [g(\{\mathbb{S}, \mathbb{A}\})] = \Sigma_\mathbb{A} + \Sigma_\mathbb{S}\\
& = \mathbf{E}_\mathbb{S}[\mathbf{Var}_\mathbb{A}[g(\{\mathbb{S}, \mathbb{A}\}|\mathbb{S})]] + \mathbf{Var}_\mathbb{S}[\mathbf{E}_\mathbb{A}[g(\{\mathbb{S}, \mathbb{A}\}|\mathbb{S})]], \end{eqnarray*}
where $\Sigma_\mathbb{A}$ denotes the variance due to stochasticity in the action selection, and $\Sigma_\mathbb{S}$ is the variance due to visiting a limited number of states before gradient estimation.  
In $g(\{\mathbb{S}, \mathbb{A}\})$,  $\pi_\mathcal{D}$ is the only part that depends on the random variable $\mathbb{A}$, which gets cancelled out when the ratio $w(s,a)$ equals to $\frac{\pi_\theta(a|s)}{\pi_\mathcal{D}(a|s)}$ in case of the estimate $g_{is}$.
This can be achieved when SR-generated functional policy accurately captures the sampling data distribution, i.e.  $\tilde{\pi}_{sym}=\pi_\mathcal{D}$.   
Therefore, $w(s,a)=\frac{\pi_\theta(a|s)}{\tilde{\pi}_{sym}(a|s)};~\tilde{\pi}_{sym}=\pi_\mathcal{D}$ indicates $\Sigma_\mathbb{A}=0$ for $g_{is}$ though $\Sigma_\mathbb{A}\neq 0$ for $g_{mc}$, which implies: $\mathbf{Var}_{\mathbb{S}, \mathbb{A}} [g_{is}(\{\mathbb{S}, \mathbb{A}\})] < \mathbf{Var}_{\mathbb{S}, \mathbb{A}} [g_{mc}(\{\mathbb{S}, \mathbb{A}\})]$.


\bibliographystyle{kr}
\bibliography{references}

\begin{thebibliography}{}

\bibitem[\protect\citeauthoryear{Alibekov, Kubalik, and
  Babuska}{2016}]{Babuska_cdc2016}
Alibekov, E.; Kubalik, J.; and Babuska, R.
\newblock 2016.
\newblock Symbolic method for deriving policy in reinforcement learning.
\newblock In {\em IEEE 55th Conference on Decision and Control (CDC)},
  2789--2795.
\newblock Las Vegas, NV, USA: IEEE.

\bibitem[\protect\citeauthoryear{Arulkumaran \bgroup et al\mbox.\egroup
  }{2017}]{surveyRL_ieee2017}
Arulkumaran, K.; Deisenroth, M.~P.; Brundage, M.; and Bharath, A.~A.
\newblock 2017.
\newblock Deep reinforcement learning: A brief survey.
\newblock {\em IEEE Signal Processing Magazine} 34(6):26--38.

\bibitem[\protect\citeauthoryear{Asadi \bgroup et al\mbox.\egroup
  }{2017}]{stat2017}
Asadi, K.; Allen, C.; Roderick, M.; Mohamed, A.-r.; Konidaris, G.; Littman, M.;
  and Amazon, B.~U.
\newblock 2017.
\newblock Mean actor critic.
\newblock {\em stat} 1050:1.

\bibitem[\protect\citeauthoryear{Aspis \bgroup et al\mbox.\egroup
  }{2022}]{KR2022}
Aspis, Y.; Broda, K.; Lobo, J.; and Russo, A.
\newblock 2022.
\newblock Embed2sym-scalable neuro-symbolic reasoning via clustered embeddings.
\newblock In {\em International Conference on Principles of Knowledge
  Representation and Reasoning}, volume~19,  421--431.

\bibitem[\protect\citeauthoryear{Biggio \bgroup et al\mbox.\egroup
  }{2021}]{transformerSR_icml2021}
Biggio, L.; Bendinelli, T.; Neitz, A.; Lucchi, A.; and Parascandolo, G.
\newblock 2021.
\newblock Neural symbolic regression that scales.
\newblock In {\em International Conference on Machine Learning},  936--945.
\newblock Virtual: PMLR.

\bibitem[\protect\citeauthoryear{Duan \bgroup et al\mbox.\egroup
  }{2016}]{benchmark2016}
Duan, Y.; Chen, X.; Houthooft, R.; Schulman, J.; and Abbeel, P.
\newblock 2016.
\newblock Benchmarking deep reinforcement learning for continuous control.
\newblock In {\em International conference on machine learning},  1329--1338.
\newblock PMLR.

\bibitem[\protect\citeauthoryear{Ferreira \bgroup et al\mbox.\egroup
  }{2022}]{1KR2022}
Ferreira, J.; de~Sousa~Ribeiro, M.; Gonçalves, R.; and Leite, J.
\newblock 2022.
\newblock Looking inside the black-box: Logic-based explanations for neural
  networks.
\newblock In {\em International Conference on Principles of Knowledge
  Representation and Reasoning}, volume~19,  432--442.

\bibitem[\protect\citeauthoryear{Gao, Schulman, and Hilton}{2022}]{chatGPT2022}
Gao, L.; Schulman, J.; and Hilton, J.
\newblock 2022.
\newblock Scaling laws for reward model overoptimization.
\newblock {\em arXiv} 2210.10760:1--28.

\bibitem[\protect\citeauthoryear{Green, Vineyard, and Koc}{2018}]{Green2018}
Green, S.; Vineyard, C.~M.; and Koc, C.~K.
\newblock 2018.
\newblock Impacts of mathematical optimizations on reinforcement learning
  policy performance.
\newblock In {\em International Joint Conference on Neural Networks (IJCNN)},
  1--8.
\newblock IEEE.

\bibitem[\protect\citeauthoryear{Hanna and Stone}{2019}]{aamas_2019}
Hanna, J., and Stone, P.
\newblock 2019.
\newblock Reducing sampling error in the monte carlo policy gradient estimator.
\newblock In {\em Proceedings of the 18th International Conference on
  Autonomous Agents and Multiagent Systems}.
\newblock Montreal: IFAAMAS.

\bibitem[\protect\citeauthoryear{Hein, Udluft, and Runkler}{2018}]{Hein2018}
Hein, D.; Udluft, S.; and Runkler, T.~A.
\newblock 2018.
\newblock Interpretable policies for reinforcement learning by genetic
  programming.
\newblock {\em Engineering Applications of Artificial Intelligence}
  76:158--169.

\bibitem[\protect\citeauthoryear{Isayev}{2018}]{ReLeaSE_Github}
Isayev, O.
\newblock 2018.
\newblock Release (reinforcement learning for structural evolution).
\newblock \url{https://github.com/isayev/ReLeaSE}.

\bibitem[\protect\citeauthoryear{Jie and Abbeel}{2010}]{nips_2010}
Jie, T., and Abbeel, P.
\newblock 2010.
\newblock On a connection between importance sampling and the likelihood ratio
  policy gradient.
\newblock In {\em Advances in Neural Information Processing Systems},
  volume~23.
\newblock NIPS.

\bibitem[\protect\citeauthoryear{Kim, Kim, and
  Petersen}{2021}]{guiSR_ijcai2021}
Kim, J.~T.; Kim, S.~K.; and Petersen, B.~K.
\newblock 2021.
\newblock An interactive visualization platform for deep symbolic regression.
\newblock In {\em Proceedings of the Twenty-Ninth International Conference on
  International Joint Conferences on Artificial Intelligence},  5261--5263.
\newblock Montreal-themed virtual reality: IJCAI.

\bibitem[\protect\citeauthoryear{Kubalik \bgroup et al\mbox.\egroup
  }{2021}]{Babuska_ieeeAccs2021}
Kubalik, J.; Derner, E.; Zegklitz, J.; and Babuska, R.
\newblock 2021.
\newblock Symbolic regression methods for reinforcement learning.
\newblock {\em IEEE Access} 9:139697--139711.

\bibitem[\protect\citeauthoryear{Landajuela \bgroup et al\mbox.\egroup
  }{2021}]{SymbPolicy_icml2021}
Landajuela, M.; Petersen, B.~K.; Kim, S.; Santiago, C.~P.; Glatt, R.; Mundhenk,
  N.; Pettit, J.~F.; and Faissol, D.
\newblock 2021.
\newblock Discovering symbolic policies with deep reinforcement learning.
\newblock In {\em International Conference on Machine Learning},  5979--5989.
\newblock Virtual: PMLR.

\bibitem[\protect\citeauthoryear{Levine and Koltun}{2013}]{offPolicyIS_2013}
Levine, S., and Koltun, V.
\newblock 2013.
\newblock Guided policy search.
\newblock In {\em International conference on machine learning},  1--9.
\newblock PMLR.

\bibitem[\protect\citeauthoryear{Popova, Isayev, and
  Tropsha}{2018}]{denovoRL_Science2018}
Popova, M.; Isayev, O.; and Tropsha, A.
\newblock 2018.
\newblock Deep reinforcement learning for de novo drug design.
\newblock {\em Science Advances} 4(7):eaap7885.

\bibitem[\protect\citeauthoryear{Rogers and Hahn}{2010}]{fpMorgan_2010}
Rogers, D., and Hahn, M.
\newblock 2010.
\newblock Extended-connectivity fingerprints.
\newblock {\em Journal of Chemical Information and Modeling} 50(5):742--754.

\bibitem[\protect\citeauthoryear{Silver \bgroup et al\mbox.\egroup
  }{2018}]{gamesRL_deepMind_Science2018}
Silver, D.; Hubert, T.; Schrittwieser, J.; Antonoglou, I.; Lai, M.; Guez, A.;
  and Lanctot, M. e.~a.
\newblock 2018.
\newblock A general reinforcement learning algorithm that masters chess, shogi,
  and go through self-play.
\newblock {\em Science} 362(6419):1140--1144.

\bibitem[\protect\citeauthoryear{Soni \bgroup et al\mbox.\egroup
  }{2022}]{rsos_2022}
Soni, T.; Sharma, A.; Dutta, R.; Dutta, A.; Jayavelu, S.; and Sarkar, S.
\newblock 2022.
\newblock Capturing functional relations in fluid–structure interaction via
  machine learning.
\newblock {\em Royal Society open science} 9(4):220097.

\bibitem[\protect\citeauthoryear{Stephens}{2018}]{gplearn_2018}
Stephens, T.
\newblock 2018.
\newblock gplearn: Genetic programming in python.

\bibitem[\protect\citeauthoryear{Sutton \bgroup et al\mbox.\egroup
  }{1999}]{SuttonPG_nips1999}
Sutton, R.~S.; McAllester, D.; Singh, S.; and Mansour, Y.
\newblock 1999.
\newblock Policy gradient methods for reinforcement learning with function
  approximation.
\newblock In {\em Advances in neural information processing systems},
  volume~12.
\newblock Morgan Kaufmann Publishers Inc.

\bibitem[\protect\citeauthoryear{Szepesvari}{2010}]{lecture_algosRL}
Szepesvari, C.
\newblock 2010.
\newblock Algorithms for reinforcement learning.
\newblock {\em Synthesis lectures on artificial intelligence and machine
  learning} 4(1):1--103.

\bibitem[\protect\citeauthoryear{Tokdar and Kass}{2010}]{impSample_review2010}
Tokdar, S.~T., and Kass, R.~E.
\newblock 2010.
\newblock Importance sampling: a review.
\newblock {\em Wiley Interdisciplinary Reviews: Computational Statistics}
  2(1):54--60.

\bibitem[\protect\citeauthoryear{Udrescu and Tegmark}{2020}]{AIFeynman_2020}
Udrescu, S.-M., and Tegmark, M.
\newblock 2020.
\newblock Ai feynman: A physics-inspired method for symbolic regression.
\newblock {\em Science Advances} 6(16):eaay2631.

\bibitem[\protect\citeauthoryear{Williams}{1992}]{REINFORCE1992}
Williams, R.~J.
\newblock 1992.
\newblock Simple statistical gradient-following algorithms for connectionist
  reinforcement learning.
\newblock {\em Machine learning} 8:229--256.

\end{thebibliography}

\end{document}